\documentclass[10pt, a4paper]{article}
\usepackage{lrec2000}
\usepackage{alltt, epsfig, url}

\usepackage{boxedminipage}
\usepackage{latexsym}
\newenvironment{sv}{\scriptsize\begin{alltt}}{\end{alltt}\normalsize}
\def\smtt#1{{\small\tt #1}}
\def\boxfig#1{\fbox{\parbox{0.97\linewidth}{\centerline{\epsfig{#1}}}}}
\def\myurl#1{{\small [\url{#1}]}}
\def\arXivhack{\vspace{-6pt}}

\title{Models and Tools for Collaborative Annotation}

\name{Xiaoyi Ma, Haejoong Lee, Steven Bird and Kazuaki Maeda} 

\address{
  Linguistic Data Consortium, University of Pennsylvania\\
  3615 Market Street, Philadelphia, PA 19104-2608, USA\\
\{xma, haejoong, sb, maeda\}@ldc.upenn.edu
}

\abstract{The Annotation Graph Toolkit (AGTK) is a collection of software
which facilitates development of linguistic annotation tools. AGTK
provides a database interface which allows applications to use a
database server for persistent storage.  This paper discusses various
modes of collaborative annotation and how they can be supported with
tools built using AGTK and its database interface. We describe the
relational database schema and API, and describe a version of the
TableTrans tool which supports collaborative annotation.  The
remainder of the paper discusses a high-level query language for annotation graphs,
along with optimizations, in support of expressive and efficient access to the
annotations held on a large central server.
The paper demonstrates that it is straightforward to support a variety of
different levels of collaborative annotation with existing AGTK-based
tools, with a minimum of additional programming effort.
}

\begin{document}

\maketitleabstract

\section{Introduction}

Annotation graphs provide a comprehensive formal framework for constructing,
maintaining and searching linguistic annotations, while remaining
consistent with many alternative data structures and file formats. An
annotation graph is a directed acyclic graph where edges are labeled
with fielded records, and nodes are (optionally) labeled with time
offsets. The annotation graph model is capable of representing
virtually all types of linguistic annotation in widespread use
today \cite{BirdLiberman01}.

The Annotation Graph Toolkit (AGTK) provides software infrastructure
based on annotation graphs, allowing developers
to quickly create special-purpose annotation tools using common
components. AGTK consists of three parts: the annotation graph library,
which is the internal data structure of annotation graphs, the I/O library,
which handles the input and output of files of different formats, and
wrappers which provide interfaces for scripting languages Tcl and Python.

AGTK also provides a model and tools to facilitate collaborative
annotation. In typical annotation projects, tasks are divided into multiple
passes; different people work on different passes individually or
together. At any given time, only one person edits an annotation file.
With AGTK it is straightforward to develop tools that permit a
group of people (who may be geographically dispersed) to collaborate on the same
annotation project.  AGTK provides a database interface which allows users
to load or retrieve annotation graphs from a shared server.  The database
interface is flexible with respect to the database server software and with
respect to the location of the data.

This paper is organized as follows.  Section 2 describes various kinds
of collaborative annotation, while section 3 lays out the database
schema and the related API functions. Section 4 is a case study of
real world problem and how it was addressed with the model we
propose. Section 5 describes our efforts towards more expressive and
efficient annotation graph queries, and section 6 concludes the paper.
\arXivhack

\section{Collaborative Annotation}

In recent years, annotation projects have grown in size and complexity,
and it is now standard for multiple annotators and sites to
be involved in a single large project.  Managing this collaboration becomes
a significant task in its own right.

At a minimum, we would like to be able to control access to different
regions and types of annotation, to log modifications, and track the
quality checks that have been made.  While version control software
and database servers may meet these requirements, it is not trivial to
incorporate this functionality into annotation tools and expose it to
end-users.  Instead, we seek lightweight solutions that can easily be
integrated with existing tools.

In this section we describe three approaches to this problem that we are
pursuing in the context of the Annotation Graph Toolkit.  These are
treated in order of increasing difficulty.

\subsection{Exploiting the annotations}

Annotation graphs permit labels to contain fielded records.
There is constraint on the name and content of the fields, except that they
must be strings.  There can be an arbitrary number of fields.
We can exploit this flexibility in managing an annotation project
by storing management information with the annotations themselves.
Thus, there could be fields for such things as the identity of the
last person who edited the annotation, various significant dates in
the lifetime of the annotation, the level of quality control
which has been completed, the chapter and verse of the coding manual which
justifies the annotation, free text comments about additional checks that
should be undertaken, and so forth.

API functions that provide search capabilities over the annotation
label data permit exactly the same functionality for this management
information.  Tools can hide this information, but use it in
presenting annotation content to users.  For instance, annotations may
have a QC feature whose values ranges from 1 to 5.  For a particular
task, the tool could highlight all annotations with QC level less than
3.

While this management information resides in the live database, it
does not need to live in all (or any) exported formats.  When
saving to particular formats, the appropriate export module only
queries the annotation graph library for the relevant fields; all
other fields are ignored.  Thus, annotation graphs can be
enriched with management information in unforseen ways, yet saved in
the existing supported formats without any modification to the export
modules.

This feature of annotation graphs is amenable to collaborative
annotation, since it is easy for the cooperating parties to agree on
additional fields which help document and manage their joint work.  We
will not have anything further to say about this approach here, but
want to emphasize that it is adequate for managing many kinds of
collaborative annotation.

\subsection{Exploiting the database}

The database access method provided by the annotation graph library uses
the ODBC standard (Open Database Connectivity).  In the past, if a single
program needed to connect to an Oracle database, an Informix database or a
MySQL database, it was necessary to maintain
three versions of the interface code, one for each database.  With ODBC,
applications write to the ODBC API and let the ODBC Manager and Driver
take care of the database language specifics.

The annotation graph database schema, which is explained in detail in
next section, allows annotation graphs to be stored and retrieved in
any relational database server that supports ODBC. The annotations can
then be queried directly in SQL or in a customized query language.  A
client-side annotation tool can initiate queries and display
annotation content on behalf of the end user. An annotation tool and
server, integrated using the model shown below, enables users to
access local or remote annotation databases transparently.

\begin{figure}[htbp]
  \begin{center}
    \leavevmode
    \epsfig{figure=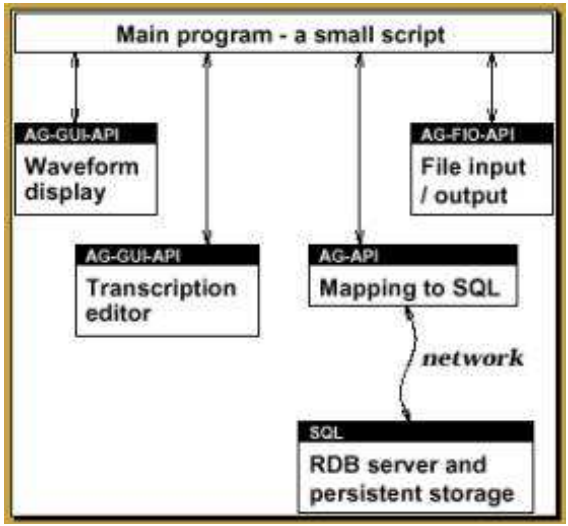,width=0.9\linewidth}
  \end{center}
  \caption{Interactions Among Annotation Tools and the Annotation Server}
  \label{fig:model}
\end{figure}

Incorporating ODBC support in AGTK greatly facilitates data sharing
and collaborative annotation.  Different users may be granted
different levels of access to the server, e.g.  to modify existing
annotations, or just to add commentary to existing annotations
(e.g. write a prose recommendation that a certain annotation be
changed).

Annotation graph based annotation tools can also be tailored to permit
certain categories of user to do certain kinds of edits.  This
supports annotation projects where there is a high degree of
specialization amongst the geographically separated collaborators
(e.g. if we were annotating discourse and intonation, and if junior
annotators were given simple tasks, and senior annotators and
researchers added their own specialized annotations, and checked the
work of the junior annotators).

Updates may occur at various levels of granularity. For example, with
record-level locking, two annotators could be working on the same
annotation region without interfering with each other's work. Section
4 also gives an example of how annotators can collaborate using column
locking where different annotators are given different read/write
permissions to edit certain annotation features.

So far we have focussed on collaborative annotation as the key benefit
derived from moving to a relational database model. There are several
other advantages as well.

With ODBC, an annotation application can access remote annotation
servers.  The user simply tells the application what annotation graph
database on which server he/she wants to access by using an ODBC
connect string which contains information of the name of the server,
the name of the database, the user name and password.  Note that the
signal data does not need to reside in the same place as the
annotation data.  Distributed annotators may have local copies of
large signal files while storing their annotations in a common central
location.

While other annotation tools are able to store their annotations in a
database \cite{Cassidy99}, we are not aware of any projects which use
this to support collaborative annotation.

We briefly note one final, significant benefit of storing all
annotations in relational form.  As Cieri and Bird
\newcite{CieriBird01} have discussed, many annotated recordings come
with a variety of non-temporal data, such as speaker demographics,
lexicons, and the like.  These tables can be stored in relational
form alongside the annotations, enabling us to integrate all of
the data.  Now, queries can perform joins across the temporal
and atemporal data.  For further discussion the reader is referred
to \cite{CieriBird01}.

\subsection{Exploiting the query language}

In many annotation projects, the only way to view an annotation is with the
same software that was used to create it.  For any large annotation task --
especially those involving collaboration -- browsing individual annotations
is usually not an efficient way to identify annotations requiring
further work.  The standard solution is to create special-purpose scripts
which scan a collection of annotation files, opening the editor on each
file which satisfies the user's requirements.

Storing the annotations in a database is an obvious win since SQL
queries can be used to efficiently identify the annotations requiring
further attention.  Such queries may even operate across multiple
corpora.  Unfortunately, SQL has some expressive limitations which
render it unsuitable for certain kinds of query.  For instance, we
cannot express kleene closure over the annotation relation, since this
requires a variable number of joins.  However, regular expressions
are a common feature of linguistic queries, and are heavily used in
the special-purpose scripts (mentioned above).

In section 5 we will report the result of our experiments on pre-compiling
the kleene closure so that these linguistically natural queries can be
expressed.  This work promises to greatly improve the flexibility of the
annotation tools, permitting users to identify and load previously-created
annotations according to complex criteria without leaving the annotation tool.
\arXivhack

\section{Relational Representation}

This section discusses the database schema and the API that are used for
storing and accessing a set of annotation graphs.

\begin{figure}[htbp]
  \begin{center}
    \leavevmode
    \epsfig{figure=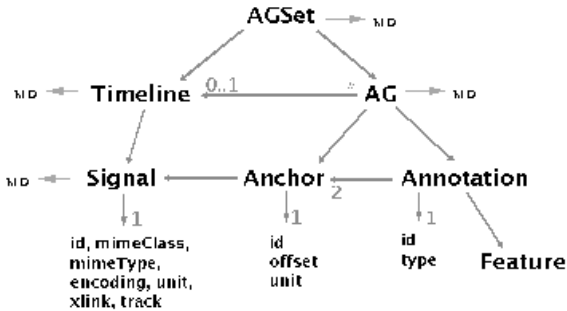, width=\linewidth}
  \end{center}

  \caption{The Annotation Graph Object Model}
  \label{fig:object-model}
\end{figure}

The design of the relational database schema is closely related to the
AG library's C++ implementation. Figure~\ref{fig:object-model} depicts
how AG library objects relate to each other. As the diagram shows, an
AGSet is a collection of timelines and AGs.  A timeline is a
collection of signals. An AG contains multiple anchors and
annotations. The annotation graph library objects also reference each
other, for example, an annotation graph can reference a
timeline; an annotation references two anchors (the start and
end anchor); a Metadata object ('MD' in the figure) could be referenced
by an AGSet, AG, timeline or signal. There are also attributes
associated with each library object, as shown in the picture. Please
see \cite{lrec-devel} for further details.

\subsection{Schema}

The model described above is represented in a relational database
as follows.

\begin{description}

\item[AGSET]

Table AGSET stores attributes of all AGSets.  XMLNS is the XML namespace
of the ATLAS Interchange Format (AIF). XLINK specifies the XML
Linking Language (XLink) specification the AGSet is using.

\begin{sv}
CREATE TABLE AGSET (
  AGSETID     VARCHAR(50) NOT NULL,
  VERSION     CHAR(10),
  XMLNS       CHAR(30),
  XLINK       CHAR(30),
  PRIMARY KEY (AGSETID))
\end{sv}

\item[AG]

Table AG stores attributes of all AGs. AGSETID specifies the AGSet an AG belongs to. TIMELINEID indicates the timeline an AG is associated with. An AG can also have an optional type, indicated by TYPE.

\begin{sv}
CREATE TABLE AG (
  AGID        VARCHAR(50) NOT NULL,
  AGSETID     VARCHAR(50) NOT NULL,
  TIMELINEID  VARCHAR(50),
  TYPE        CHAR(10),
  PRIMARY KEY (AGID),
  FOREIGN KEY (AGSETID) REFERENCES AGSET)
\end{sv}

\item[TIMELINE]

Table TIMELINE stores attributes of all timelines.

\begin{sv}
CREATE TABLE TIMELINE (
  AGSETID     VARCHAR(50) NOT NULL,
  TIMELINEID  VARCHAR(50) NOT NULL,
  PRIMARY KEY (TIMELINEID)
  FOREIGN KEY (AGSETID) REFERENCES AGSET)
\end{sv}

\item[SIGNAL]

Table SIGNAL keeps attributes of all signals. AGSETID and TIMELINEID specify
the AGSet and timeline a signal belongs to.

\begin{sv}
CREATE TABLE SIGNAL (
  AGSETID     VARCHAR(50) NOT NULL,
  TIMELINEID  VARCHAR(50) NOT NULL,
  SIGNALID    VARCHAR(50) NOT NULL,
  MIMECLASS   VARCHAR(50),
  MIMETYPE    VARCHAR(50),
  ENCODING    VARCHAR(50),
  UNIT        VARCHAR(50),
  XLINKTYPE   VARCHAR(50),
  XLINKHREF   VARCHAR(50),
  TRACK       VARCHAR(50),
  PRIMARY KEY (SIGNALID),
  FOREIGN KEY (AGSETID) REFERENCES AGSET,
  FOREIGN KEY (TIMELINEID) REFERENCES TIMELINE)
\end{sv}

\item[ANNOTATION]

Table ANNOTATION keeps attributes of all annotations. AGSETID and AGID
specifies the AGSet and the AG an annotation belongs to. STARTANCHOR
and ENDANCHOR are the IDs of the start anchor and end anchor of the
annotation. An annotation also has an type.

\begin{sv}
CREATE TABLE ANNOTATION (
  AGSETID      VARCHAR(50) NOT NULL,
  AGID         VARCHAR(50) NOT NULL,
  ANNOTATIONID VARCHAR(50) NOT NULL,
  STARTANCHOR  VARCHAR(50),
  ENDANCHOR    VARCHAR(50),
  TYPE         VARCHAR(50),
  PRIMARY KEY (ANNOTATIONID),
  FOREIGN KEY (AGID) REFERENCES AG,
  FOREIGN KEY (AGSETID) REFERENCES AGSET)
\end{sv}

\item[ANCHOR]

Table ANCHOR stores attributes of all anchors. AGSETID and AGID
specify the AGSet and the AG an anchor belongs to. OFFSET is the
offset of an anchor. UNIT specifies the unit for the offset. SIGNALS
indicates the signals that an anchor refers to.

\begin{sv}
CREATE TABLE ANCHOR (
  AGSETID     VARCHAR(50) NOT NULL,
  AGID        VARCHAR(50) NOT NULL,
  ANCHORID    VARCHAR(50) NOT NULL,
  OFFSET      FLOAT,
  UNIT        VARCHAR(50),
  SIGNALS     VARCHAR(50),
  PRIMARY KEY (ANCHORID),
  FOREIGN KEY (AGID) REFERENCES AG,
  FOREIGN KEY (AGSETID) REFERENCES AGSET)
\end{sv}

\item[METADATA]

Table METADATA stores all the metadescriptions for AGSet, AG,
Timeline and Signal. ID could be AGSETID, AGID, TIMELINEID or
SIGNALID.  These metadescriptions could use the Dublin Core
elements to identify the title, creator and date of the work,
and to give a prose description of its contents.  For more
discussion of the use of metadata for describing
language resources, see \cite{BirdSimons01}.

\begin{sv}
CREATE TABLE METADATA (
  AGSETID     VARCHAR(50) NOT NULL,
  AGID        VARCHAR(50),
  ID          VARCHAR(50) NOT NULL,
  NAME        VARCHAR(50) NOT NULL,
  VALUE       TEXT,
  PRIMARY KEY (ID,NAME),
  FOREIGN KEY (AGSETID) REFERENCES AGSET)
\end{sv}

\end{description}

Note that some values of the above tables, such as AGSETID, are stored
redundantly to enable efficient loading. For example, this permits all
the anchors of an AGSet can be loaded in a single query, which cannot
be done if AGSETID is not stored in the ANCHOR table.

\begin{description}

\item[Feature Tables]

AGTK also creates a feature table for each corpus. Each column of the
feature table represents one possible feature of an annotation. The name of
the column is the feature name and its value is the feature value.  For
example, to represent features of corpus ABC with possible feature names
GENDER, AGE and POB (place of birth), AGTK creates the following table:

\begin{sv}
CREATE TABLE ABC (
  ANNOTATIONID VARCHAR(50) NOT NULL,
  GENDER      TEXT,
  AGE         TEXT,
  POB         TEXT,
  PRIMARY KEY (ANNOTATIONID),
  FOREIGN KEY (ANNOTATIONID) REFERENCES ANNOTATION)
\end{sv}

Since an application can add or remove annotation features, AGTK must
add or remove columns in the corresponding feature table.  AGTK takes
care of this when storing data back to the server, i.e. it changes the
structure of the table if necessary before it stores annotation data
back to the feature table.

\end{description}

\subsection{AG database API}

The AG database interface provides load and store functions.

\paragraph{LoadFromDB}
\begin{sv}
void LoadFromDB(string connStr, AGSetId agsetId);
\end{sv}

\smtt{LoadFromDB} loads the specified \emph{AGSet} from the database
server to memory. The variable \smtt{connStr} specifies the connection string
that ODBC uses to connect to the server. It contains information such
as hostname, database name, user name, and password.

Table~\ref{tab:connectstr} shows some of the parameters used in a
connect string, for a complete list, see
\myurl{http://www.mysql.com/doc/M/y/MyODBC_connect_parameters.html}.

\begin{table*}[htbp]
  \begin{center}
    \leavevmode
{\footnotesize
\begin{tabular}{|l|l|} \hline

ODBC connect string arguments & What the argument specifies \\ \hline
DSN & Registered ODBC Data Source Name. \\ \hline
SERVER & The hostname of the database server. \\ \hline
UID & User name as established on the server. \\
& SQL Server this is the login name. \\ \hline
PWD & Password that corresponds with the login name.\\ \hline
DATABASE & Database to connect to. If not given, DSN is used.\\ \hline
\end{tabular}
}
  \end{center}
  \caption{Parameters in Connect String}
  \label{tab:connectstr}
\end{table*}

DSN is the registered ODBC Data Source Name.  It should be defined in
the .odbc.ini file in user's home  directory. All other arguments can be
either defined in the .odbc.ini file, or defined in the connect string
itself.
To gain access to most ODBC data sources, the user must provide a valid
user ID and corresponding password. These values are initially
registered by the database administrator.
The following is a sample driver section
for DSN ``talkbank'' in the configuration file for iODBC. To make
explanation easier, line numbers are included.  Note that UID
and PWD become USER and PASSWORD in iODBC's configuration file.

\begin{sv}
1  [talkbank]
2  Driver   = /usr/local/lib/libmyodbc.so
3  DSN      = talkbank
4  SERVER   = talkbank.ldc.upenn.edu
5  USER     = myuserid
6  PASSWORD = mypasswd
7  DATABASE = talkbank
\end{sv}

Line 1 is the name of the driver section, ``talkbank'' (the user can
have multiple driver sections in a single configuration file). Line 2
specifies which ODBC driver to use. Line 3 gives the name of the DSN,
which is ``talkbank''. Line 4 specifies the hostname of the machine on
which the database server is running. Lines 5 and 6 give the user name
and password for connecting to the server.  Line 7 identifies the
database server to connect to.  With a complete .odbc.ini file such as
the above, the connection string can be just \smtt{DSN=talkbank;}.  If
the user has not specified some of the arguments in the configuration
file, say USER and PASSWORD, he/she can still specify them in the
connect string:

\begin{sv}
DSN=talkbank;UID=myuserid;PWD=mypasswd;
\end{sv}

\paragraph{StoreToDB}

\begin{sv}
void StoreToDB(string connStr, AGSetId agSetId);
\end{sv}

\smtt{StoreToDB} stores the specified AGSet to the database
server. The variable \smtt{connStr} contains connection information
as explained above.
\arXivhack

\section{Collaborative Annotation with TableTrans}

In this section we describe a collaborative annotation need which has been
presented to us, and how this need can be addressed using the TableTrans
tool.  The context is research by Robert Seyfarth, Dorothy Cheney and
colleagues at the University of Pennsylvania on social behavior and vocal
communication in nonhuman primates.  These researchers make extended audio
recordings of primate interactions, making detailed notes of the social
context and physical environment.  Later, they listen to the recordings and
identify particular vocalizations of interest, noting their start and end
times and classifying the call types.  Observations may be dense, or very
sparse with extended periods in which nothing is coded.  Each observation
is entered into a row of a spreadsheet, and rows may have upwards of a
dozen columns, each covering a different aspect of the observation, e.g.:
recording offsets, tape number, date, time, location, animal id, group id,
context (foraging/predator), call type, signal quality, and comments.  Each row
may also contain quantities derived from the corresponding period of the
signal, such as mean energy.  In this way, many thousands of observations
are coded.  Quantitative analysis of these tables addresses research
questions in behavioral ecology and language evolution.

Much of the coding task is relatively straightforward and does not
require highly trained annotators.  Working from a digitized recording
and field notes, an annotator can convert tape counter numbers to
millisecond offsets, and enter such fields as the tape number, date,
time and location.  This work is typically done at a digitization
station; while tapes are being uploaded to disk, the annotator works
on previously uploaded materials.  The result is a set of spreadsheets
in which each row corresponds to an extent of audio, but where certain
columns are left empty.

In the original version of this process, annotation files would remain
on the digitization station until the first round of annotation was
complete, at which time they would be transferred to a specialist --
usually the original field researcher -- for further annotation.  The
specialist would fill in columns that require a greater degree of
critical judgement, such as the call type.  However, during the course
of listening to hundreds of calls, the untrained annotator gradually
learns to discriminate most call types, and can usefully make a first
pass at annotating some of the specialized columns.  Later, these can
be reviewed and post-edited by the specialist.

\begin{figure}[htbp]
  \begin{center}
    \leavevmode
    \epsfig{figure=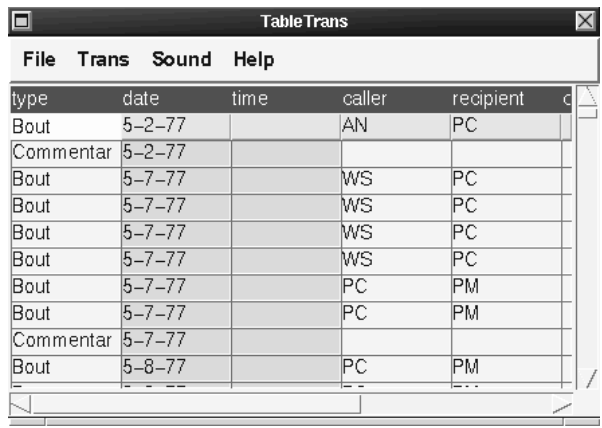, width=0.95\linewidth}
  \end{center}
  \caption{TableTrans With Read-Only Columns}
  \label{fig:TableTransColLock}
\end{figure}

The unfortunate consequence of this regime is that the collaboration
between the annotator and the specialist requires copies of the
annotations to be circulated (e.g. by email) and/or arranging
meetings.  Neither of these is an efficient way to quickly resolve the
unpredictable questions that may arise (and hold up) the annotation
process.  
In addressing this problem, we have added a new capability to the
TableTrans tool \cite{lrec-demo} which enables the user to choose to store
all of its annotations in a central database. 
The configurations for the trainee and specialist versions
of the tool specify which columns are available for read-only access,
and which columns are available for read-write access.  Even if
physically separated, both parties can review the same material,
listen to the audio segments, and discuss judgements (e.g. by email or
telephone).  In this way, collaboration, training and quality control
can be done remotely, while each person has full access to the
up-to-date annotations.  The tool is able to prevent unauthorized
modification of read-only columns, and indicates these columns
by shading (see Figure~\ref{fig:TableTransColLock}).

This mode of collaborative annotation is possible because the data resides
on a shared server, and because the tool can enforce a simple policy for
the permission to edit particular spreadsheet columns.
\arXivhack

\section{Efficient AG Queries}

Given that annotation graphs are stored in a relational database, it seems
reasonable to use SQL for search.  However, SQL turns out to be unsuitable
for annotation graphs in terms of expressiveness and optimizability.  Thus,
a new query language for annotation graphs has been proposed
\cite{BirdBunemanTan00}.  In most cases, the new query language can be
mapped into SQL.  Therefore, it is still possible to utilize SQL and
relational database technologies.  There are a couple of advantages in
this approach.  First, given that we are taking full advantage of a
relational database system, implementation is straightforward; we only
need to map between query languages.  Second, the optimization problem
is reduced to the optimization of SQL queries, in which there is
relatively little room for improvement.

The major problem in mapping between the proposed annotation graph
query language and SQL is the arbitrary steps of arc tracing which
cannot be expressed well in SQL.  In this section, we propose a
solution for the problem.  Also, with a series of experiments, we
probe the feasibility of this mapping approach in terms of efficiency
which is critical in real time applications.

\subsection{The experimental database}

For the experiment, we set up a test database on a Linux PC running at
500 MHz.  PostgreSQL \cite{PostgreSQL} was used for the relational
database.  For test data, a part of the TIMIT corpus \cite{TIMIT86}
was used, consisting of 1,680 TIMIT speech annotations.
The contents of the database are summarized in the following table:\footnote{
  The test database can be tuned up for better performance.  For instance,
  some additional indices can be added, and management queries can be used
  for clean-up.  We used a tuned database in the experiments reported here.
}

\begin{center}
\begin{tabular}{|l|r|}
\hline
\multicolumn{1}{|c|}{Object} & \multicolumn{1}{c|}{Count} \\
\hline\hline
Agset & 1 \\
\hline
Ag & 1,680 \\
\hline
Annotation & 80,378 \\
\multicolumn{1}{|r|}{\emph{txt}} & \multicolumn{1}{r|}{1,680} \\
\multicolumn{1}{|r|}{\emph{wrd}} & \multicolumn{1}{r|}{14,553} \\
\multicolumn{1}{|r|}{\emph{phn}} & \multicolumn{1}{r|}{64,145} \\
\hline
Anchor & 67,375 \\
\hline
\end{tabular}
\end{center}

\subsection{AG query example}

Consider the following annotation graph query against our test database:

\begin{enumerate}
\item[] \textbf{Query 1.} Find word arcs whose phonetic transcription starts with a `hv' and contains a `dcl'.
\end{enumerate}

\begin{figure}[h]
\boxfig{figure=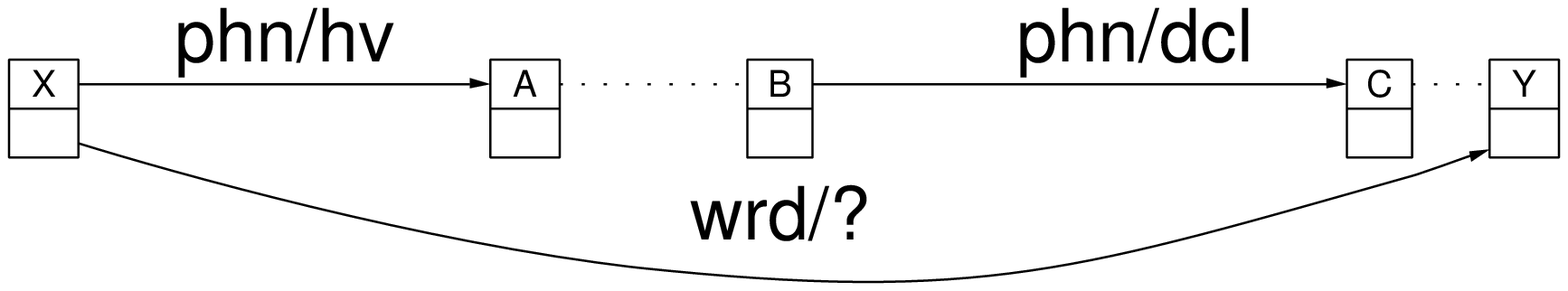, width=0.9\linewidth}
\caption{Annotation graph pattern for query 1.}
\label{fig:ag_pattern}
\end{figure}

This query can be depicted by an annotation graph pattern as shown in
Figure~\ref{fig:ag_pattern}.  It matches any \emph{wrd} annotation
that starts at anchor $X$ and ends at anchor $Y$ such that there exist
a \emph{phn} annotation labeled `hv' that starts at anchor $X$ and
ends at anchor $A$, a \emph{phn} annotation labeled `dcl' that starts
at anchor $B$ and ends at anchor $C$, paths $A \leadsto B$ and $C
\leadsto Y$ of \emph{phn} annotations.

This pattern can be written as follows using annotation graph query syntax:

\begin{sv}
    X.[].Y <- db/wrd
    X.[:hv].[]*.[:dcl].[]*.Y <- db/phn
\end{sv}

\noindent
Here is a complete annotation graph query for query 1.

\begin{sv}
    SELECT I
    WHERE  X.[id:I].Y <- db/wrd AND
           X.[:hv].[]*.[:dcl].[]*.Y <- db/phn;
\end{sv} 

Note that \smtt{id:I} is added to the pattern to select ids of matched
annotations.  Also note that there is no direct mapping to SQL for the
above query because of the paths of \emph{phn} annotations.  The
solution for this problem is discussed in the following sections.

\subsection{$K^*$: Transitive closure of annotations}

$K^*$ is a table containing connectability information between
anchors.  For example, if there is a path from anchor $A$ to anchor
$B$ following annotations of type $t$, $K^*$ should contain a tuple
$(A,B,t)$, and vice versa.  The schema for $K^*$ is shown below:

\begin{sv}
CREATE TABLE Kstar (
  StartAnchor  VARCHAR(50),
  EndAnchor    VARCHAR(50),
  Type         VARCHAR(20),
  PRIMARY KEY (StartAnchor,EndAnchor,Type),
  FOREIGN KEY (StartAnchor,EndAnchor) REFERENCES Anchor);
\end{sv} 

Precomputing $K^*$, the transitive
closure of annotations, eliminates the mapping problem.  In fact,
query 1 is now translated to SQL in Figure~\ref{fig:query1.sql}.

\begin{figure}[t]
\begin{boxedminipage}[t]{\linewidth}
\begin{sv}
SELECT W.annotationid

FROM  (SELECT AnnotationId, StartAnchor, EndAnchor
       FROM   Annotation
       WHERE  Type='wrd') AS W,

      (SELECT StartAnchor, EndAnchor
       FROM   Annotation A, TIMIT F
       WHERE  A.Type='phn' AND
              A.AnnotationId=F.annotationId AND
              F.Label='hv') AS P1,

      (SELECT StartAnchor, EndAnchor
       FROM   Annotation A, TIMIT F
       WHERE  A.Type='phn' AND
              A.AnnotationId=F.annotationId AND
              F.Label='dcl') AS P2,

      (SELECT StartAnchor, EndAnchor
       FROM   Kstar
       WHERE  Type='phn') AS K1,

      (SELECT StartAnchor, EndAnchor
       FROM   Kstar
       WHERE  Type='phn') AS K2

WHERE  W.StartAnchor=P1.StartAnchor AND
       P1.EndAnchor=K1.StartAnchor AND
       K1.EndAnchor=P2.StartAnchor AND
       P2.EndAnchor=K2.StartAnchor AND
       K2.EndAnchor=W.EndAnchor
;
\end{sv}
\end{boxedminipage}
\caption{SQL query for query 1.}
\label{fig:query1.sql}
\end{figure}

\paragraph{Domain Restriction}

In the query in Figure~\ref{fig:query1.sql}, $K^*$ tuples of
type \emph{phn} are used.  This is wasteful because it allows
irrelevant anchors to be considered in the computation.  For example, for
two connected anchors, the SQL query will take those anchors into
consideration as long as they are connected by \emph{phn}
annotations, even though they belong to different words.

To avoid this problem, we can restrict the domain of computation of
transitive closure.  In the above case, we can compute the transitive
closure only for the \emph{phn} annotations that belong to the same
word.  This will make the size of $K^*$ much smaller, making queries
faster.  The following table shows statistics for $K^*$ for the test
database.  The transitive closures are computed for
\emph{wrd} and \emph{phn} annotations and for \emph{phn} annotations
within the \emph{wrd} domain, denoted by \emph{phn/wrd}.

\begin{center}
\begin{tabular}{|c|r|}
\hline
\multicolumn{1}{|c|}{Type} & \multicolumn{1}{c|}{Count} \\
\hline\hline
\emph{wrd} & 91,960 \\
\hline
\emph{phn} & 1,423,134 \\
\hline
\emph{phn/wrd} & 271,323 \\
\hline\hline
Total & 1,786,417 \\
\hline
\end{tabular}
\end{center}

The following queries are variations of query 1.

\begin{enumerate}
\item[] \textbf{Query 2.} Find word arcs whose phonetic transcription starts with a `hv':
\vspace{-2mm}
\begin{sv}
SELECT I
WHERE  X.[id:I].Y <- db/wrd
       X.[:hv].[]*.Y <- db/phn;
\end{sv}

\item[] \textbf{Query 3.} Find word arcs whose phonetic transcription starts with a `hv' and ends with a `ix':
\vspace{-2mm}
\begin{sv}
SELECT I
WHERE  X.[id:I].Y <- db/wrd
       X.[:hv].[]*.[:ix].Y <- db/phn;
\end{sv}

\item[] \textbf{Query 4.} Find word arcs whose phonetic transcription contains a `dcl':
\vspace{-2mm}
\begin{sv}
SELECT I
WHERE  X.[id:I].Y <- db/wrd
       X.[]*.[:dcl].[]*.Y <- db/phn;
\end{sv}
\end{enumerate}

The following table shows the performance of those queries and
the effect of domain restriction.

\begin{center}
\begin{tabular}{|c|r|r|}
\hline
Query & \multicolumn{1}{c|}{\emph{phn}} & \multicolumn{1}{c|}{\emph{phn/wrd}} \\
\hline\hline
1 & 2.22 & 1.59 \\
\hline
2 & 0.85 & 0.79 \\
\hline
3 & 2.40 & 2.41 \\
\hline
4 & 22.70 & 3.90 \\
\hline
\end{tabular}
\end{center}

Since $K^*$ is a huge table, it is expected that queries with many
references to $K*$ will suffer from the size of $K^*$.  In considering
this issue, suppose that we have the following general pattern for
an annotation graph query:

\begin{sv}
    X.[].Y <- db/wrd
    X.[:\emph{l1}].[]*.[:\emph{l2}].[]*......[:\emph{ln}].[]*.Y <- db/phn
\end{sv}

\begin{figure}[t]
\boxfig{figure=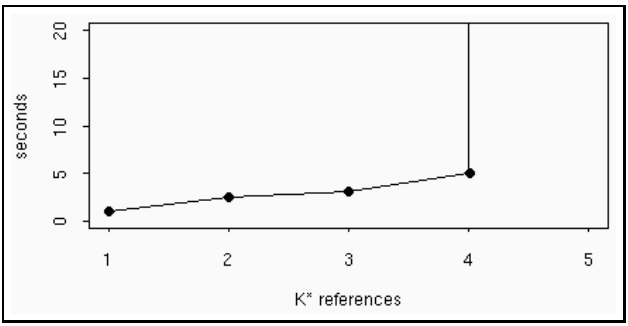, width=\linewidth}
\caption{Long joins with $K^*$.}
\label{fig:long_joins}
\end{figure}

The pattern has $n$ \emph{phn} annotations and each pair of \emph{phn} annotations has an intervening $K^*$ reference, a total of $n$ $K^*$ references.  Figure~\ref{fig:long_joins} shows how the performance scales as the number of $K^*$ references increases.  With 5 $K^*$ references, it took 23 hours and 40 minutes with a lot of memory swapping.

\subsection{$K^*-$array: Alternative to $K^*$}

$K^*$ shows poor performance when the query contains large number of
$K^*$ references.  In this section, we propose an alternative to $K^*$
to solve the problem.

Consider a boolean $n \times n$ matrix, where $n$ is the number of
anchors in an annotation graph.  We require a distinct matrix
for each type $t$.  The value of cell $(i,j)$ is
\emph{true} iff there is a path from anchor $i$ to anchor $j$
following annotations of type $t$.\footnote{Note that it is still
possible to apply domain restriction.  For example, the value of cell
$(i,j)$ is \emph{false} if both $i$ and $j$ don't belong to the same
domain.}  In general, the number of matrices required for an annotation
graph is $N_t$ (number of annotation types) + $N_d$ (number of domain
restrictions).  Therefore, if tuples have a matrix as their attribute,
the size of the $K^*$ table can be reduced to ($N_t$+$N_d$) $\times$
(number of annotation graphs).

A table is built as described above, using PostgreSQL's array data
type, and we call it $K^*-$array.  The schema is shown below:

\begin{sv}
CREATE TABLE Kstar (
  AGId         VARCHAR(50),
  A            BOOL[][],
  Type         VARCHAR(20),
  PRIMARY KEY (AGId, Type),
  FOREIGN KEY (AGId) REFERENCES AG);
\end{sv} 

The size of $K^*-$array for our
test database is 5,040; remember the size of $K^*$ is 1.8 million.
Note that $K^*-$array reduces not only the size of the table but also the
number of joins.  For instance, Figure~\ref{fig:query1a.sql} is the
$K^*-$array version of the SQL query in Figure~\ref{fig:query1.sql},
where two connectability tests are done by one $K^*-$array join.
(\smtt{anchor\_num()} is a function to map anchor ids to a natural
number.)  We can expect that this will improve performance
for queries involving a large number of joins.

\begin{figure}[t]
\begin{boxedminipage}[t]{\linewidth}
\begin{sv}
SELECT W.annotationid

FROM  (SELECT AnnotationId,StartAnchor,EndAnchor,AGId
       FROM   Annotation
       WHERE  Type='wrd') AS W,

      (SELECT StartAnchor, EndAnchor
       FROM   Annotation A, TIMIT F
       WHERE  A.Type='phn' AND
              A.AnnotationId=F.annotationId AND
              F.Label='hv') AS P1,

      (SELECT StartAnchor, EndAnchor, AGId
       FROM   Annotation A, TIMIT F
       WHERE  A.Type='phn' AND
              A.AnnotationId=F.annotationId AND
              F.Label='dcl') AS P2,

      (SELECT AGId, A
       FROM   Kstar_array
       WHERE  Type='phn') AS K

WHERE  W.StartAnchor=P1.StartAnchor AND
       K.AGId=W.AGId AND
       P2.AGId=W.AGId AND
       K.A[anchor_num(P1.EndAnchor)][anchor_num(P2.St\
artAnchor)] AND
       K.A[anchor_num(P2.EndAnchor)][anchor_num(W.End\
Anchor)]
;
\end{sv}
\end{boxedminipage}
\caption{SQL query for query 1 with $K^*-$array.}
\label{fig:query1a.sql}
\end{figure}

The following table shows the effect of the $K^*-$array for
queries 1-4.  There are clearly time savings, although there is little
benefit from domain restriction.

\begin{center}
\begin{tabular}{|c|r|r|r|r|}
\hline
& \multicolumn{2}{c|}{$K^*$} & \multicolumn{2}{c|}{$K^*-$array} \\
\cline{2-5}
Query & \emph{phn} & \emph{phn/wrd} & \emph{phn} & \emph{phn/wrd} \\
\hline\hline
1 & 2.22 & 1.59 & 1.24 & 1.24 \\
\hline
2 & 0.85 & 0.79 & 0.57 & 0.57 \\
\hline
3 & 2.40 & 2.41 & 2.40 & 2.40 \\
\hline
4 & 22.70 & 3.90 & 2.42 & 2.24 \\
\hline
\end{tabular}
\end{center}

\begin{figure}[t]
\boxfig{figure=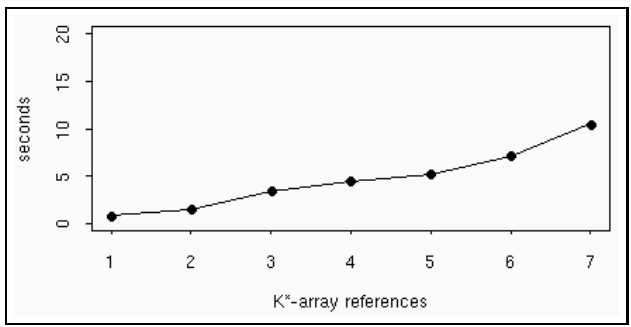, width=\linewidth}
\caption{Long joins with $K^*-$array.}
\label{fig:long_joins2}
\end{figure}

Figure~\ref{fig:long_joins2} demonstrates that the $K^*-$array can
handle queries having a large number of joins.

\subsection{Future work}

The result of these experiments shows that this approach to
implementing annotation query is quite promising.  There are, however,
remaining tasks to complete the implementation.  Tasks include: (i)
implementation of translator (or mapper) from annotation graph query
to SQL, (ii) further experimentation on vertical queries to allow
efficient search on hierarchical information between annotations such
as inclusion and overlapping, and (iii) specifying details of the
query language syntax.
\arXivhack

\section{Conclusion}

This paper has reported on models and tools for collaborative annotation
based on annotation graphs.  Annotation graphs are an extremely general and
versatile data model for representing all kinds of annotation data.
Moreover, they can be stored in a relational database and accessed
remotely.  The Annotation Graph Toolkit supports connections to
ODBC-compliant database servers, making it easy for developers to create
annotation tools that store all their data in a server.  Apart from the
obvious benefits for data management, this storage method opens up new
possibilities for collaborative annotation, as reported above.  We hope to
have shown that it is completely straightforward to support a variety of
different levels of collaborative annotation with existing AGTK-based
tools, with a minimum of additional programming effort.
\arXivhack

\section*{Acknowledgements}

This material is based upon work supported by the National Science
Foundation under Grant Nos. 9978056 and 9980009 (Talkbank).

\bibliographystyle{lrec2000k}

\end{document}